\documentclass[12pt,halfline,a4paper]{ouparticle}
\usepackage[english]{babel}
\usepackage[colorlinks=true,citecolor=blue,linkcolor=blue,urlcolor=blue,hyperfootnotes=false]{hyperref}
\usepackage{natbib}
\usepackage[utf8]{inputenc}
\usepackage{booktabs}
\usepackage{multirow}
\usepackage[dvipsnames]{xcolor}
\usepackage{marvosym}
\usepackage{fontawesome}
\usepackage[bottom]{footmisc}

\usepackage{silence}
\usepackage[T1]{fontenc}
\usepackage{upgreek}
\usepackage{breqn}

\usepackage{rotating}

\begin{document}

\title{Emulating Public Opinion: A Proof-of-Concept of AI-Generated Synthetic Survey Responses for the Chilean Case\thanks{This working paper was funded by Empiria Lab and Pular Consulting. We thank the Pular Consulting team for their feedback during this work. The views expressed and any remaining errors are the authors' own.}}

\author{%
\name{Bastián González-Bustamante\thanks{Associate Professor, Faculty of Administration and Economics, Universidad Diego Portales, Chile, {\large \Letter} \href{mailto:bastian.gonzalez.b@mail.udp.cl}{bastian.gonzalez.b@mail.udp.cl}. Postdoctoral Researcher in Computational Social Science, Faculty of Governance and Global Affairs, Leiden University, Netherlands, {\large \Letter} \href{mailto:b.a.gonzalez.bustamante@fgga.leidenuniv.nl}{b.a.gonzalez.bustamante@fgga.leidenuniv.nl}. ORCID iD \href{https://orcid.org/0000-0003-1510-6820}{https://orcid.org/0000-0003-1510-6820}, {\faLinkedinSquare} \href{https://www.linkedin.com/in/bgonzalezbustamante}{bgonzalezbustamante}, {\faHome} \href{https://bgonzalezbustamante.com/}{https://bgonzalezbustamante.com}.}}
\address{Universidad Diego Portales \\ Leiden University}
\and
\name{Nando Verelst\thanks{PhD Researcher, Faculty of Administration and Economics, Universidad Diego Portales, Chile, {\large \Letter} \href{mailto:nando.verelst@mail.udp.cl}{nando.verelst@mail.udp.cl}. ORCID iD \href{https://orcid.org/0000-0002-0957-767X}{https://orcid.org/0000-0002-0957-767X}, {\faLinkedinSquare} \href{https://www.linkedin.com/in/nando-verelst-402b9711a/}{nando-verelst-402b9711a}.}}
\address{Universidad Diego Portales}
\and
\name{Carla Cisternas\thanks{PhD Researcher, Faculty of Humanities, Leiden University, Netherlands, {\large \Letter} \href{mailto:c.g.cisternas.guasch@hum.leidenuniv.nl}{c.g.cisternas.guasch@hum.leidenuniv.nl}. ORCID iD \href{https://orcid.org/0000-0001-7948-6194}{https://orcid.org/0000-0001-7948-6194}, {\faLinkedinSquare} \href{https://www.linkedin.com/in/carlacisternasguasch/}{carlacisternasguasch}, {\faHome} \href{https://carlacisternas.com/}{https://carlacisternas.com}.}}
\address{Leiden University}
}

\abstract{Large Language Models (LLMs) offer promising avenues for methodological and applied innovations in survey research by using synthetic respondents to emulate human answers and behaviour, potentially mitigating measurement and representation errors. However, the extent to which LLMs recover aggregate item distributions}

\date{}


\maketitle

{\quotation \noindent \small remains uncertain and downstream applications risk reproducing social stereotypes and biases inherited from training data. We evaluate the reliability of LLM-generated synthetic survey responses against ground-truth human responses from a Chilean public opinion probabilistic survey. Specifically, we benchmark 128 prompt-model-question triplets, generating 189,696 synthetic profiles, and pool performance metrics (i.e., accuracy, precision, recall, and F1-score) in a meta-analysis across 128 question-subsample pairs to test for biases along key sociodemographic dimensions. The evaluation spans OpenAI’s GPT family and o-series reasoning models, as well as Llama and Qwen checkpoints. Three results stand out. First, synthetic responses achieve excellent performance on trust items (F1-score and accuracy $> 0.90$). Second, GPT-4o, GPT-4o-mini and Llama 4 Maverick perform comparably on this task. Third, synthetic-human alignment is highest among respondents aged 45-59. Overall, LLM-based synthetic samples approximate responses from a probabilistic sample, though with substantial item-level heterogeneity. Capturing the full nuance of public opinion remains challenging and requires careful calibration and additional distributional tests to ensure algorithmic fidelity and reduce errors. \endquotation}

\begin{center}
\end{center}

\section{Introduction}

Traditional public opinion surveys face a number of challenges and risks related to measurement and representation dimensions, including, for example, coverage error due to incomplete frames and hard-to-reach groups, sampling error resulting from finite samples and complex designs, nonresponse error stemming from low participation and interview fatigue, measurement error introduced by questionnaire wording, and processing errors in coding and post-survey adjustments, among others \citep{Groves1989, Groves2010, Weisberg2005}. These errors could be amplified by substantial financial, human, and logistical demands, such as time spent on instrument design, piloting, and fieldwork that often forces a cost-quality trade-off that may distort population inferences. Consequently, there is a growing demand in the social sciences and market research for methods that reduce burden and cost while maintaining and improving overall data quality.

Against this backdrop, Large Language Models (LLMs), trained extensively on vast and diverse data, emerge as promising alternatives for new research possibilities and applied research, including handling the abovementioned survey research limitations and measurement and representation errors. Indeed, recent advances in generative artificial intelligence (AI) suggest LLMs could serve for a number of classification tasks, including the creation of synthetic samples, providing simulated responses reflective of broader societal attitudes and behaviours \citep{Argyle2023, Gilardi2023, GonzalezBustamante2024}. The synthetic samples specifically may leverage the ability of LLMs to generate contextually informed responses based on individual-level demographic characteristics and attitudes, and, in this way, potentially emulate public opinion without direct interaction with human respondents. This methodological innovation opens new avenues for rapid data collection, experimentation with sensitive topics, and a deeper understanding of complex public opinion dynamics that complement or even partially substitute for traditional surveys.

Thus, the primary objective of this working paper is to evaluate the effectiveness and reliability of LLM-generated synthetic survey responses in reflecting real-world public opinion in Chile. Specifically, we aim to assess the predictive accuracy of a number of state-of-the-art private and open-source LLMs by comparing their synthetic respondents against human probabilistic responses.\footnote{By a synthetic respondent, we mean an AI-generated persona used to simulate a survey participant. We create these personas by giving an LLM basic traits, as we further detail in the method section. Then, we ask it to answer as if it were that person.} Furthermore, we examine the impact of incorporating a variety of contextual information, including demographic and attitudinal variables, on the predictive performance of the models. In doing so, we contribute to the understanding of the feasibility and potential downstream applications of synthetic survey methodologies and LLMs in public opinion research.

\section{From Synthetic Data to Synthetic Publics: LLMs in Public Opinion Research}

Synthetic data have a long tradition in the social sciences, initially used to deal with nonresponse using multiple imputation (\citeauthor{Rubin1993}, \citeyear{Rubin1993}; see also \citeauthor{Little1993}, \citeyear{Little1993}; \citeauthor{Little1987}, \citeyear{Little1987}; \citeauthor{Raghunathan2003}, \citeyear{Raghunathan2003}). However, while the current research in synthetic data is mainly motivated by harvesting considerable data to train machine learning models, some government agencies are exploring these approaches to broaden public access to their resources \citep{Drechsler2023}. Indeed, there is potential related to privacy-preserving and cost-efficient substitutes for microdata, aiming to reproduce joint distributions without exposing individual, sensitive records. Although there are several applications, from patient information and health records \citep{Choi2018, Yahi2017} to environmental monitoring \citep{Allken2019}, all of them share a common goal: the synthetic data must be similar to the results obtained using actual data \citep{Drechsler2023}. In this sense, synthetic data may not be merely a way to deal with missingness or privacy concerns for survey-oriented applications; when properly constructed and validated, they can act as stand-ins for hard-to-obtain samples, enabling faster iteration and lower-cost implementation.

A number of early applications originated from techniques developed to address class imbalance and small-{\itshape N} constraints. Methods like Synthetic Minority Oversampling Technique (SMOTE) and its variants showed that augmenting minority classes can reduce the accuracy paradox\footnote{Accuracy, as a classic machine learning indicator, reflects the proportion of correct predictions of a model in the test set, however, it could be misled by data imbalance because the model is predicting the majority class and is not able to identify underrepresented categories \citep{GonzalezBustamante2023, GonzalezBustamante2025b}.} and improve sensitivity in rare event settings.\footnote{The foundational work in synthetic sampling, notably the SMOTE, was indeed primarily designed to mitigate class imbalance by generating examples for the minority class. SMOTE remains effective nowadays. For instance, \cite{deBlasio2022} demonstrated that it was important for transforming a model with zero sensitivity for detecting rare instances of municipal corruption into a viable tool for policy intervention. Without synthetic oversampling, their model failed to identify positive cases, a classic example of the accuracy paradox. Similarly, \cite{Walid2022} and \cite{Tchokote2025} found that advanced SMOTE variants were effective for enhancing model robustness and prediction accuracy in educational data mining and multimodal hate speech detection tasks, respectively.} Subsequent research generalised this intuition to broader data scarcity, showing that generators which preserve covariance structure and local dependencies (e.g., optimised multivariable Kernel Density Estimation, KDE, for small biomedical datasets or fuzzy diffusions via {\itshape k}-nearest neighbours) can expand datasets while maintaining multivariate relationships \citep{Fowler2020, Sivakumar2022}. For survey research, the relevance of these advances lies less in their original domains than in the transferable principle that credible synthetic data must respect the joint distribution of key covariates rather than merely reproduce marginals.

Modern methods have been increasingly tackling the challenge of data scarcity and sparsity, moving beyond the focus on class distribution. \cite{Fowler2020}, for example, developed a method using optimised multivariate KDE to generate entire synthetic populations from small biomedical datasets. Their objective was not to balance classes but to prevent overfitting by meticulously replicating the complex, correlated covariance of the original data, thereby creating a sufficiently large and representative dataset for modelling where none existed before. This focus on the small data problem is also evident in the work of \cite{Sivakumar2022}, who created the kNN Mega-Trend Diffusion (kNNMTD) method. This approach is agnostic to the learning task, whether regression or classification, and is designed explicitly for tiny datasets. Indeed, the kNNMDT method is able to generate plausible synthetic data that faithfully maintains the original data correlation structure by applying a fuzzy set-based information diffusion technique locally via {\itshape k}-nearest neighbours, preventing overestimation.

Parallel advances in deep generative modelling, such as Generative Adversarial Networks (GANs; see \citeauthor{Goodfellow2014}, \citeyear{Goodfellow2014}) and diffusion models, extended synthetic data beyond tabular settings to sequential and textual modalities.\footnote{Moving beyond simple tabular data, for example, some techniques could generate synthetic instances in sparse regions of the feature space and assign a new label distribution by interpolating the ones of neighbouring instances \citep{Gonzalez2021}. Other progress has been made in the integration of synthetic sampling and multimodal pipelines \citep[see][]{Tchokote2025}.} Beyond their use in time-series \citep[see][]{deSouza2023} and text \citep[see][]{Tolba2021}, GANs have been effectively applied to augment structured survey data in the social sciences. \cite{RuizGandara2025} demonstrated that GANs could generate synthetic survey responses that maintained the original data’s statistical behaviour, spurring the development of novel harmonic mean-based indices for quality evaluation. In addition, comparative studies have emphasised that quality matters more than quantity \citep{deSouza2023}, proposing evaluation frameworks that benchmark not only predictive accuracy, but also distributional similarity based on measures such as Maximum Mean Discrepancy (MMD) and PCA-based (i.e., based on Principal Component Analysis) diagnostics \citep{Fowler2020}, along with correlation-preservation metrics such as Pairwise Correlation Difference (PCD; see \citeauthor{Sivakumar2022}, \citeyear{Sivakumar2022}).\footnote{Alternative options for exploring the distribution are measures such as Earth-Mover’s Distance (EMD; see \citeauthor{Boelaert2025}, \citeyear{Boelaert2025}; \citeauthor{GonzalezBustamante2019}, \citeyear{GonzalezBustamante2019}; \citeauthor{Lupu2017}, \citeyear{Lupu2017}).}

Against this methodological backdrop, a new strand evaluates LLMs directly as synthetic respondents. The core design emulates survey interviews by conditioning an LLM on respondent profiles based on demographics and, when available, attitudinal traits \citep{Argyle2023, Kim2024}. Evidence is mixed but promising: LLMs can recover aggregate distributions for specific items, reproduce well-known covariates-attitude associations, and display sensitivity to profile granularity and prompt wording \citep{Argyle2023, Bisbee2024}. \cite{Shrestha2024} explored the use of GPT-4 to generate synthetic data for policy surveys across diverse cultural contexts, reaching a reasonable aggregate alignment with human responses, however, they also uncovered systematic positive biases and cultural variability in the accuracy of synthetic data. \cite{Boelaert2025} indicated, on the other hand, that current models could hardly emulate human respondents because of machine bias. In sum, this mixed evidence highlights both the potential of LLMs to create large-scale synthetic data and the inherent risk of introducing biases that may be difficult to detect because of the intrinsic patterns from pretraining data and instruction tuning datasets \citep{Geng2024, GonzalezBustamante2024}. 

Indeed, the risk for downstream applications based on LLMs could be associated with encoding social stereotypes or political biases, yielding subgroups distortions or directional tilt on contentious issues \citep{Morris2025, Santurkar2023, Qu2024}. For this reason, the level of algorithmic fidelity, which reflects how LLMs can replicate effectively socio-cultural contexts and nuanced viewpoints of subpopulations, is quite relevant \citep{Argyle2023, Ma2025}. In this sense, safeguarding algorithmic fidelity should motivate evaluation practices that move beyond headline accuracy by focusing on subgroups, evaluation of prompt perturbations \citep[see][]{Bisbee2024}, among other innovative approaches. For survey emulation, validating the synthetic distribution against probabilistic samples should be considered as a starting point to test the quality of the AI-generated synthetic profiles.

\section{Methodology}

\subsection{Survey Emulation and Probabilistic Sample}

For survey emulation and validation of the synthetic distribution, we rely on the probabilistic sample survey conducted by the Centro de Estudios Públicos (CEP) in Chile, widely recognised for its rigorous methodology and comprehensive coverage of socio-demographic and attitudinal variables. The CEP survey programme has served since 1986 as a barometer of public opinion on political and economic views, key societal concerns, and citizens’ evaluations of government performance and other major actors.

The target population of the CEP survey comprises residents aged 18 and older across Chile, excluding provinces classified as difficult to access by the National Statistics Institute (INE in Spanish), namely Easter Island, Palena, and Antarctica. The sample design is probabilistic, stratified by region and by urban-rural area, with random selection at three stages: block, household, and respondent. The instrument is a structured questionnaire incorporating programmed skip patterns, administered face-to-face using electronic devices such as tablets or smartphones to enhance accuracy and consistency in data collection. In addition, survey results are weighted and adjusted post-survey to account for selection probabilities and nonresponse, thereby correcting under- or over-representation of specific groups.

Specifically, we employed the probabilistic survey number 92, carried out between August 2 and September 12, 2024.\footnote{We created the synthetic profiles by the end of April 2025. By that time, CEP number 92 was the latest released survey. Further information is available on \href{https://www.cepchile.cl/opinion-publica}{https://www.cepchile.cl/opinion-publica}.} This survey comprised $1,482$ face-to-face interviews conducted across $127$ Chilean municipalities, yielding a response rate of $61.3\%$ from an initial sample of $2,416$. The sample error was $\pm 2.8$ with a $95\%$ confidence interval. 

\subsection{LLMs, Prompting Strategies and Performance}

We tested the performance of a number of state-of-the-art private and open-source LLMs to answer the question. Our proof-of-concept included flagship OpenAI GPTs such as GPT-4o (2024-11-20), GPT-4o-mini (2024-07-18), GPT-4.5-preview (2025-02-27), GPT-4.1, 4.1-mini and 4.1-nano (2025-04-14), as well as reasoning models like o1-mini (2024-09-12) and o3-mini (2025-01-31). We also tested the latest Llama 4 Maverick (400B) and Scout (107B), Llama 3.3 (70B), Qwen 2.5 (32B) and Gemma 3 (12B).\footnote{We tried to consider o1 (2024-12-17) and o4-mini (2025-04-16), however, being reasoning models, the tasks required considerable time. For future applications, Qwen 3 (0.6B, 1.7B, 4B, 8B, 14B, 30B, 32B and 235B), released the last week of April 2025, may be a suitable alternative. In addition, more recent models, such as GPT-5 or the novel, very first open-source OpenAI model GPT-OSS (20B and 120B), released in August 2025, should be good options. It is relevant to note that there are a variety of state-of-the-art closed and open-source flagship models developed by Antrophic, Bespoke Labs, DeepSeek-AI, Mistral, among other providers, that are worthy of trying \cite[see][]{GonzalezBustamante2025a}.}

We employed two distinct prompting strategies to generate synthetic responses. The first strategy utilises exclusively demographic information (i.e., region, urban or rural area, age, sex, educational level, socioeconomic group, and occupation). The second incorporates attitudinal dimensions, such as individuals’ political interests, ideological identification, and previous demographic characteristics. This dual approach enables us to investigate the effect of varying contextual information levels on the accuracy of LLMs. Therefore, we created 1,482 synthetic profiles per prompting strategy, which implies 38,532 synthetic profiles (1,482 respondent profiles $\times$ 2 prompting strategies $\times$ 13 LLMs) for the proof-of-concept, focusing on the question: {\itshape “According to the following scale, how would you rate the country's current economic situation? Very bad, bad, neither good nor bad, good, very good.”} 

We then benchmarked the LLMs’ predictive performance by estimating standard evaluation metrics, including accuracy, precision, recall, and F1-score \citep{Biecek2021, GonzalezBustamante2023, GonzalezBustamante2025b}. We reduced the list of models for benchmarking based on the proof-of-concept indicators\footnote{The goodness-of-prediction indicators of the proof-of-concept against economic perception did not show the best performance, however, as we will discuss in the results for the final benchmark, some model $\times$ question pairs show relatively high performance with F1-scores above 0.9, particularly for questions related to trust. This suggests that there is considerable heterogeneity not only depending on the model but also in relation to the question.} provided in the \nameref{Appendix}  in order to conduct an error-rate analysis using the questions in \textcolor{blue}{Table} \ref{TAB1} as ground-truth. Consequently, we tested GPT-4o (2024-11-20), GPT-4o-mini (2024-07-18), Llama 4 Maverick (400B), and Qwen 2.5 (32B), which involved the creation of 189,696 synthetic profiles (1,482 respondent profiles $\times$ 2 prompting strategies $\times$ 4 LLMs $\times$ 16 ground-truth questions, including the proof-of-concept question and the expanded ones for benchmark).\footnote{By a ground-truth question, we mean the human responses to a specific survey item from our reference probabilistic sample. We use this observed distribution as the benchmark to evaluate LLM-generated answers. This ground-truth is an empirical measure, not normative, and is subject to the survey sampling error indicated above.} This implies that we have 128 observations for the prompting-LLM-question triplet.

\begin{table}[!ht] \centering
\caption{Expanded Ground-Truth Questions for Benchmark}
\label{TAB1} 
\resizebox{0.9\width}{!}{%
\begin{tabular}{@{}ll@{}}
\\[-1.8ex]\hline \\[-1.8ex] 
\\[-1.8ex] Topic & Question \\ [1ex]
\hline \\[-1.8ex] 
\multirow{2}{*}{Economic perception} & Do you think that in the next 12 months the economic situation in \\
& the country will improve, stay the same or get worse? \\ \midrule
\multirow{3}{*}{Economic perception} & According to the following scale, how would you rate your current \\
& economic situation? Very bad, bad, neither good nor bad, good, \\
& very good. \\ \midrule
\multirow{2}{*}{Political perception} & How would you rate the current political situation in Chile? Very \\
& bad, bad, neither good nor bad, good, very good. \\ \midrule
\multirow{3}{*}{Government approval} & Regardless of your political position, do you approve or disapprove \\
& of the way President Gabriel Boric is leading the government? \\
&  Approves, disapproves, neither approves nor disapproves. \\ \midrule
\multirow{4}{*}{Democracy} & Which of the following statements do you most agree with? \\
& Democracy is always preferable, sometimes an authoritarian \\ 
& regime can be preferable, a democratic regime is the same as an \\
& authoritarian regime. \\ \midrule
\multirow{3}{*}{Trust} & Can you tell me how much confidence you have in the Catholic \\
& Church? A lot of confidence, fairly confident, low confidence, no \\
& confidence. \\ \midrule
\multirow{3}{*}{Trust} & Can you tell me how much confidence you have in the Armed \\
& Forces? A lot of confidence, fairly confident, low confidence, no \\
& confidence. \\ \midrule
\multirow{3}{*}{Trust} & Can you tell me how much confidence you have in the Courts of \\ 
& Justice? A lot of confidence, fairly confident, low confidence, no \\
& confidence. \\ \midrule
\multirow{3}{*}{Trust} & Can you tell me how much confidence you have in the written \\ 
& press? A lot of confidence, fairly confident, low confidence, no \\
& confidence. \\ \midrule
\multirow{2}{*}{Trust} & Can you tell me how much confidence you have in the TV? A lot  \\
& of confidence, fairly confident, low confidence, no confidence. \\ \midrule
\multirow{3}{*}{Trust} & Can you tell me how much confidence you have in the police \\
& (Carabineros de Chile)? A lot of confidence, fairly confident, low \\
& confidence, no confidence. \\ \midrule
\multirow{2}{*}{Trust} & Can you tell me how much confidence you have in the government? \\
& A lot of confidence, fairly confident, low confidence, no confidence. \\ \midrule
\multirow{3}{*}{Trust} & Can you tell me how much confidence you have in the political \\
& parties? A lot of confidence, fairly confident, low confidence, no \\
& confidence. \\ \midrule
\multirow{2}{*}{Trust} & Can you tell me how much confidence you have in Congress? A lot \\
& of confidence, fairly confident, low confidence, no confidence. \\ \midrule
\multirow{2}{*}{Trust} & Can you tell me how much confidence you have in the radio? A lot \\
& of confidence, fairly confident, low confidence, no confidence. \\
\bottomrule\\[-1.8ex] 
\end{tabular}
}\\
\end{table} 

\subsection{Meta-Analysis of Bias}

We focus GPT-4o-mini (2024-07-18) on 16 ground-truth questions crossed with eight sociodemographic subsamples to analyse potential bias in 128 combinations for the question-subsample pairs (16 questions $\times$ 8 subsamples). The subsets were elaborated based on a number of relevant sociodemographic variables, including area (urban vs. rural), gender (female vs. male), and age groups (i.e., 18-29, 30-44, 45-59, and 60 and older). We focused on the F1-score, which is appropriate given that the items' responses can be imbalanced, and then fit random-effects meta-regressions to separate sampling error from between-synthetic heterogeneity.

Our meta-analysis proceeds incrementally. The first model only incorporated a binary variable for urban area, and then we incorporated the female variable, using male as the reference category. The third model incorporated the age groups using the younger group as a reference category. In all models, we include ground-truth questions as fixed effects, which absorb item-specific baseline differences (i.e., wording and context effects). Formally, the equation for a question-subsample pair $(q, s)$ is as follows, considering $G = 16$ as the number of questions benchmarked:

\begin{dmath} 
    F1_{[q,s]} = \alpha + \beta_{1} \, urban\text{-}area_{s} + \beta_{2} \, female_{s} + \sum^{K}_{k = 2}{\gamma_{k} \, age\text{-}group_{[k,s]}} \\ + \sum^{G}_{g = 2}{\delta_{g} \, ground\text{-}truth_{[g,q]}} + u_{[q,s]} + \upvarepsilon_{[q,s]}
\end{dmath}

\section{Results}

\subsection{Error-Rate Analysis}

\begin{figure}[!ht]
\caption{Goodness-of-Prediction Metrics of Synthetic Samples Against Ground-Truth Probabilistic Sample}
\label{FIG1}
\centering \vspace{2mm}
\includegraphics[width=0.99\linewidth]{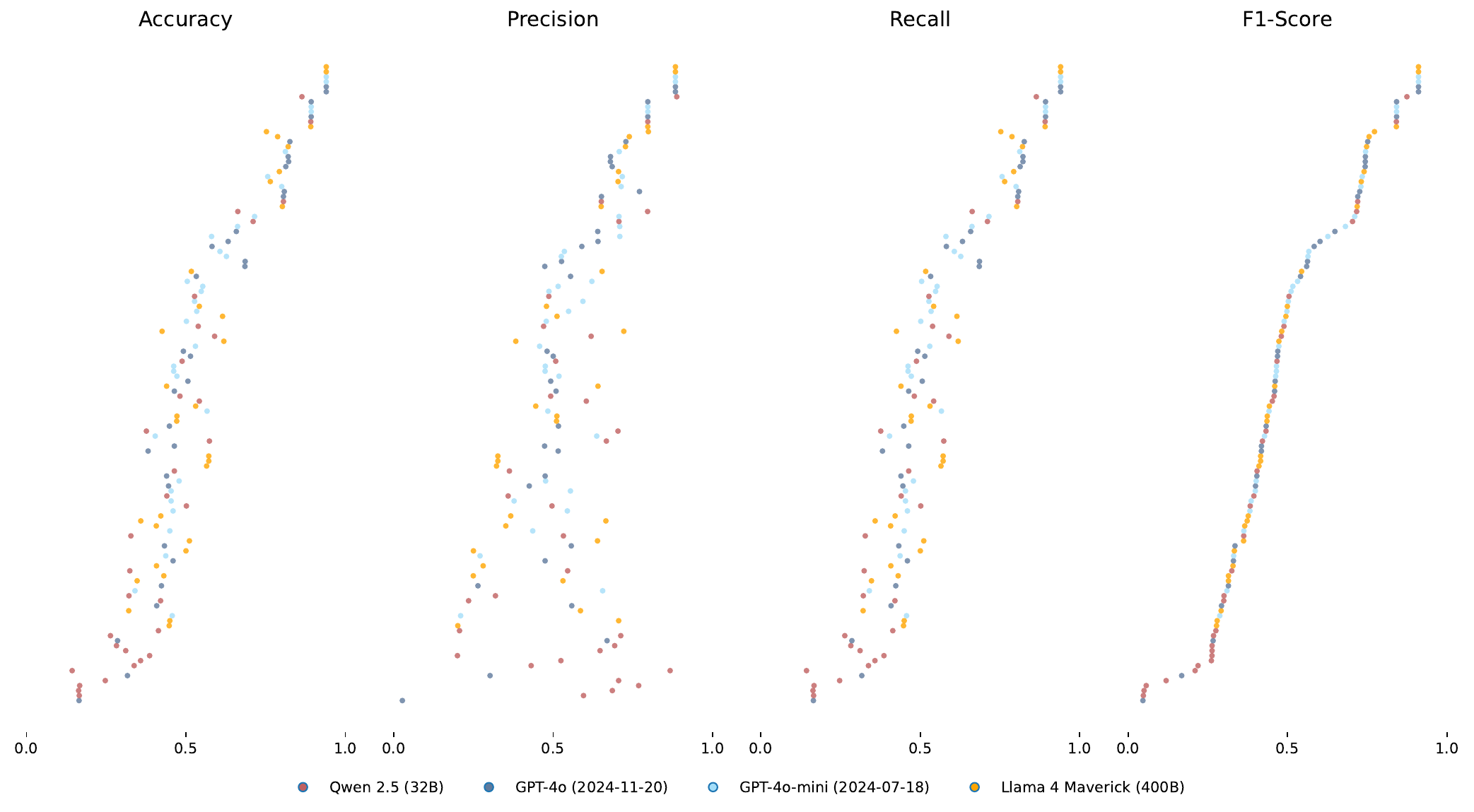} 
{\footnotesize {\itshape Note.} Accuracy reports the proportion of synthetic samples aligned with the probabilistic reference sample. Precision represents the fraction of predicted positives that are correct, and recall indicates the proportion of actual positives identified. The F1-score is the harmonic measure that combines both precision and recall.}
\end{figure}

We not only estimated the standard evaluation metrics (i.e., accuracy, precision, recall, and F1-score) summarised in \textcolor{blue}{Figure} \ref{FIG1}, but also the arithmetic mean (PPI) and the geometric mean (GPI) based on the evaluation metrics. PPI provides an intuitive sense of overall quality across different performance metrics, while GPI balances across metrics by penalising weakness in one of the metrics. These metrics provide a comprehensive assessment of each model’s ability to emulate actual respondents accurately and range from 0.696 to 0.919 for the top 15 model-question pairs available in the \nameref{Appendix}, which is a significant improvement in comparison with the original proof-of-concept.

\begin{figure}[!ht]
\caption{Accuracy and F1-Score Across Models}
\label{FIG2}
\centering \vspace{2mm}
\includegraphics[width=0.49\linewidth]{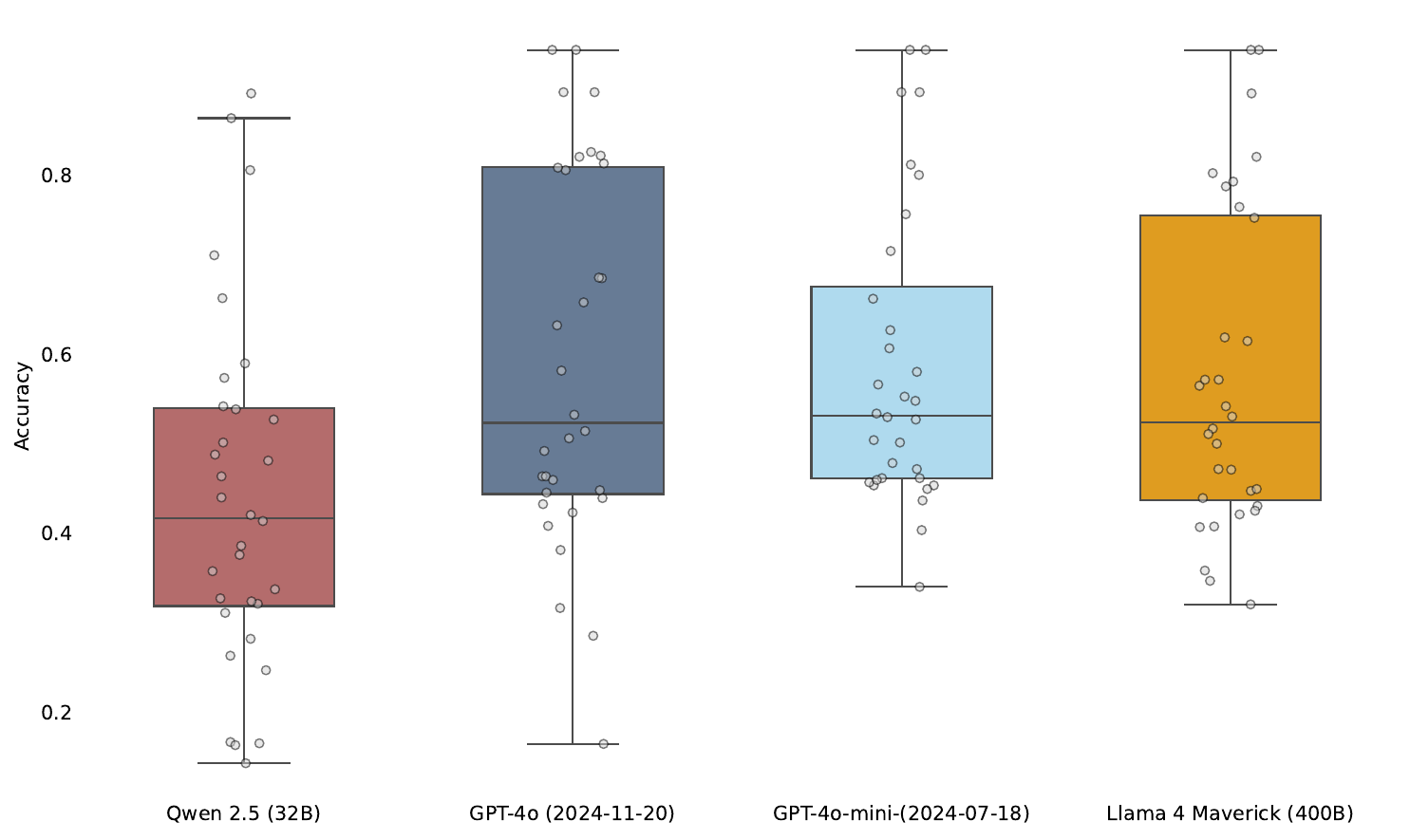} \includegraphics[width=0.49\linewidth]{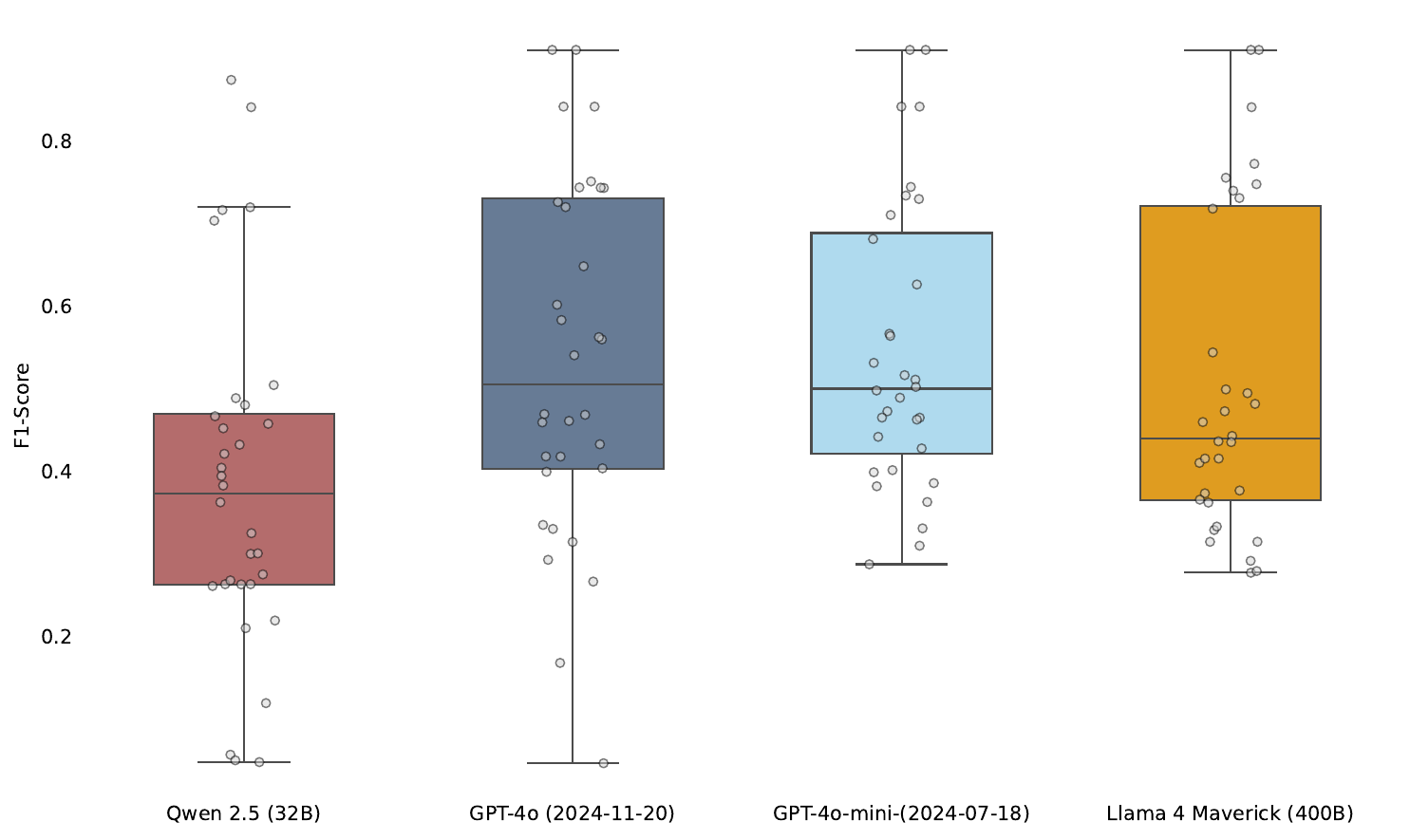} \\
{\footnotesize {\itshape Note.} Accuracy reports the proportion of synthetic samples aligned with the probabilistic reference sample. The F1-score, on the other hand, combines precision (i.e., fraction of predicted positives that are correct) and recall (i.e., proportion of actual positives identified).}
\end{figure}

Two general patterns emerge from the evidence. First, the synthetic responses on trust demonstrated excellent performance, with F1-scores and accuracy even above $0.90$. Second, the performance of GPT-4o (2024-11-20) and GPT-4o-mini (2024-07-18) is quite similar and comparable to Llama 4 Maverick (400B). \textcolor{blue}{Figure} \ref{FIG2} presents boxplots grouped by models for accuracy and F1-score, which confirm the advantages of these models in comparison with Qwen 2.5 (32B).

\subsection{Bias Analysis}

\textcolor{blue}{Table} \ref{TAB2} presents the results of our random-effects meta-regressions using the 128 valid question-subsample pairs and the logit F1-score as the dependent variable. We only found a positive, significant coefficient for the age group from 45 to 59 years old (OR = $1.145$, $\gamma = 0.136$, $95\%$ CI $[0.056, 0.215]$, $p \leq 0.001$), resulting in a $3.4$ percentage points average improvement in the F1-score in comparison with the younger group. This evidence suggests that the synthetic sample is closer to human respondents in this group. In addition, the absence of significant negative coefficients could be interpreted as a sign of the absence of bias, however, this should require further exploration of the distributions using, for example, EMD or other similar measures \citep{Boelaert2025}.

\begin{table}[!ht] \centering 
\caption{Mixed-Effects Regressions of F1-Scores from Synthetic Samples Against Ground-Truth Probabilistic Sample} 
\label{TAB2} 
\begin{tabular}{@{\extracolsep{5pt}}lccc} 
\\[-1.8ex]\hline \\[-1.8ex] 
\\[-1.8ex] & Model I & Model II & Model III \\ [1ex]
\hline \\[-1.8ex] 
\multicolumn{4}{c}{\footnotesize Reference category: Rural area} \\ \midrule
\multirow{2}{*}{Urban area} & 0.025 & 0.019 & 0.054 \\
& (0.029) & (0.030) & (0.006) \\ \midrule
\multicolumn{4}{c}{\footnotesize Reference category: Male} \\ \midrule
\multirow{2}{*}{Female} & & $-$0.030 & 0.006 \\ 
& & (0.031) & (0.035) \\ \midrule
\multicolumn{4}{c}{\footnotesize Reference category: Age 18-29} \\ \midrule
\multirow{2}{*}{Age 30-44} & & & 0.047 \\ 
& & & (0.041) \\ 
\multirow{2}{*}{Age 45-59} & & &  0.136$^{\star\star\star}$ \\ 
& & & (0.041) \\ 
\multirow{2}{*}{Age 60 or older} & & & 0.020 \\ 
& & & (0.038) \\ \midrule
\multirow{2}{*}{Constant} & 0.055 & 0.061 & 0.026 \\ 
& (0.045) & (0.045) & (0.047) \\ \midrule
Ground-truth FE & Yes & Yes & Yes \\ \midrule
$N$ & 128 & 128 & 128 \\ \midrule
\multirow{2}{*}{$\tau^{2}$} & 0.007 & 0.007 & 0.007 \\
& (0.002) & (0.002) & (0.002) \\
$\tau$ & 0.084 & 0.084 & 0.081 \\
$I^{2}$ & 44.47\% & 44.22\% & 42.45\% \\
$R^{2}$ & 0.989 & 0.989 & 0.990 \\ \midrule
AIC & $-$68.589 & $-$65.022 & $-$64.099 \\
BIC & $-$19.817 & $-$13.713 & $-$5.297 \\
\bottomrule\\[-1.8ex] 
\end{tabular} 
\\
{\footnotesize {\itshape Note.} Table entries report beta coefficient estimates and standard errors in parentheses. AIC: Akaike information criterion; BIC: Bayesian information criterion; FE: fixed-effects. \\ $^{\star}$ $p \leq 0.1$; $^{\star\star}$ $p \leq 0.05$; $^{\star\star\star}$ $p \leq 0.01$.} 
\end{table}
\vspace{2mm}

The moderators explained $99\%$ of the between-synthetic heterogeneity, and there is a low residual heterogeneity ($\tau^{2} = 0.007$, $I^{2} = 42.45\%$, in model III). This is because the ground-truth questions fixed effects have a significant impact on the models, especially those related to trust variables against the country’s current economic situation question as a reference category.\footnote{This is the very same question tested in the proof-of-concept as ground-truth.} For example, trust in political parties (OR = $9.509$, $\delta = 2.252$, $95\%$ CI $[2.104, 2.401]$, $p \leq 0.001$) and Congress (OR $= 4.961$, $\delta = 1.602$, $95\%$ CI $[1.468, 1.735]$, $p \leq$ 0.001) boosted the odds of alignment between synthetic and human responses by $40$ and $33$ percentage points in F1-score, respectively. On the contrary, Churchillian democracy (OR $= 0.466$, $\delta = -0.764$, $95\%$ CI $[-0.887, -0.641]$, $p \leq 0.001$) and prospective evaluation of the country’s economic situation (OR $= 0.589$, $\delta = -0.530$, $95\%$ CI $[-0.651, 0.409]$, $p \leq 0.001$) perform worse than the reference category by $18$ and $13$ percentage points. 

\section{Discussion and Limitations}

Taking into account our error-rate and bias analyses as evidence to evaluate the similarity of the synthetic sample generated using LLMs compared to actual human survey responses from a probabilistic sample, some takeaways stand out. First, the best models, namely GPT-4o-mini (2024-07-18) and Llama 4 Maverick (400B),\footnote{GPT-4o (2024-11-20) is similar but more expensive.} achieved an acceptable level of performance, indicating limited realism and a need for further improvement. Therefore, these synthetic respondents are able to emulate broad patterns and trends, demonstrating the potential of synthetic sampling as a cost-effective and scalable tool for exploring general attitudes and benchmarking against traditional surveys. Second, though we did not find strong evidence of bias based on sociodemographic differences in the synthetic respondents, we did not test educational or socioeconomic level, which should be evaluated in future iterations of this work.

As limitations, synthetic responses still struggle to capture the nuanced complexity of public opinion, particularly among underrepresented demographic groups, where attempts to approximate minority views can lead to larger errors \citep{Morris2025, Santurkar2023}. This outcome is unsurprising and is linked to the complexity of public opinion dynamics. It also relates to the limitations of training data, which may have restricted exposure to local contexts, cultural nuances, and linguistic variations, thereby introducing biases in the representation of specific regions or populations \citep{Qu2024}. Moreover, most of the variables we used to elaborate our synthetic profiles could interact in complex, non-linear ways. 

Although synthetic samples offer clear advantages in terms of cost, speed, and experimental flexibility, they lack the probabilistic design of survey sampling \citep{Bisbee2024}. Without careful calibration, there is a risk of exaggerating or underrepresenting some trends, jeopardising algorithmic fidelity \citep{Argyle2023, Ma2025}. For this reason, the primary value of synthetic samples lies, for the moment, in proof-of-concept exercises, sensitivity testing, and trend and complementary analysis, which allow LLMs and surveys to upgrade their capabilities mutually through, for example, alignment or fine-tuning \citep{Kim2024}.

Considering these results, future avenues for research could involve varying levels of effort and resources, ranging from low to high. First, tuning the prompting strategy would imply a low level of further effort. Second, post-processing work, such as reweighing or calibrating synthetic samples, could be a medium-effort option that would allow reestimating the actual weights of the baseline probabilistic sample and incorporating adjustments based on bias and error-rate analysis of the LLMs. In addition, another medium-effort alternative would be testing some additional recently released models, particularly some reasoning models, however, that could be extremely time-consuming. Lastly, fine-tuning and model alignment would be an excellent alternative and may be the most promising venue, however, it is time-consuming and requires more resources.

\bibliographystyle{apalike}
\bibliography{Synthetic_Samples}
\addcontentsline{toc}{section}{References}
\pagebreak

\setcounter{table}{0}
\setcounter{figure}{0}
\setcounter{subsection}{0}
\renewcommand{\thetable}{A.\arabic{table}}
\renewcommand{\thefigure}{A.\arabic{figure}}
\renewcommand{\thesubsection}{A.\arabic{subsection}}

\section*{Appendix}
\addcontentsline{toc}{section}{Appendix}
\label{Appendix}

\begin{table}[!ht] \centering
\caption{Goodness-of-Prediction Indicators of Proof-of-Concept Against Economic Perception Ground-Truth}
\label{TABA1} 
\resizebox{0.98\width}{!}{%
\begin{tabular}{@{}llcccc@{}}
\\[-1.8ex]\hline \\[-1.8ex] 
\\[-1.8ex] Model & Prompt & Accuracy & Precision & Recall & F1-Score \\ [1ex]
\hline \\[-1.8ex] 
\multirow{2}{*}{GPT-4o-mini (2024-07-18)} & Demographics & 0.530 & 0.458 & 0.530 & 0.473 \\
& Attitudes & 0.553 & 0.516 & 0.553 & 0.517 \\  \midrule
\multirow{2}{*}{o1-mini (2024-09-12)} & Demographics & 0.515 & 0.498 & 0.515 & 0.497 \\
& Attitudes	& 0.465 & 0.472 & 0.465 & 0.462 \\ \midrule
\multirow{2}{*}{o3-mini (2025-01-31)} & Demographics	& 0.534	& 0.485 & 0.534 & 0.486 \\
& Attitudes	& 0.476 & 0.467 & 0.476 & 0.468 \\ \midrule
\multirow{2}{*}{GPT-4o (2024-11-20)} & Demographics & 0.492 & 0.481 & 0.492 & 0.470 \\
& Attitudes	& 0.439 & 0.475 & 0.439 & 0.404 \\ \midrule
\multirow{2}{*}{GPT-4.5-preview (2025-02-27)} & Demographics	& 0.543 & 0.462 & 0.543 & 0.449 \\
& Attitudes	& 0.540 & 0.471 & 0.540 & 0.467 \\ \midrule
\multirow{2}{*}{GPT-4.1 (2025-04-14)} & Demographics & 0.539 & 0.468 & 0.539 & 0.416 \\
& Attitudes	& 0.542 & 0.465 & 0.542 & 0.449 \\ \midrule 
\multirow{2}{*}{GPT-4.1-mini (2025-04-14)} & Demographics & 0.540 & 0.468 & 0.540 & 0.431 \\
& Attitudes	& 0.540 & 0.473 & 0.540 & 0.435 \\ \midrule
\multirow{2}{*}{GPT-4.1-nano (2025-04-14)} & Demographics & 0.471 & 0.491 & 0.471 & 0.477 \\
& Attitudes	 & 0.509 & 0.519 & 0.509 & 0.506 \\ \midrule
\multirow{2}{*}{Llama 3.3 (70B)} & Demographics & 0.533 & 0.462 & 0.533 & 0.476 \\
& Attitudes	& 0.510 & 0.476 & 0.510 & 0.489 \\ \midrule
\multirow{2}{*}{Llama 4 Scout (107B)} & Demographics	& 0.527 & 0.478 & 0.527 & 0.444 \\
& Attitudes	& 0.512 & 0.495 & 0.512 & 0.491 \\ \midrule
\multirow{2}{*}{Llama 4 Maverick (400B)} & Demographics & 0.530 & 0.445 & 0.530 & 0.443 \\
& Attitudes	& 0.542 & 0.479 & 0.542 & 0.499 \\ \midrule
\multirow{2}{*}{Qwen 2.5 (32B)} & Demographics & 0.538 & 0.470 & 0.538 & 0.489 \\
& Attitudes	& 0.527 & 0.486 & 0.527 & 0.505 \\ \midrule
\multirow{2}{*}{Gemma 3 (12B)} & Demographics & 0.540 & 0.495 & 0.540 & 0.394 \\
& Attitudes	& 0.544 & 0.472 & 0.544 & 0.457 \\
\bottomrule\\[-1.8ex] 
\end{tabular}
}\\
{\footnotesize {\itshape Note.} Accuracy reports the proportion of synthetic samples aligned with the probabilistic reference sample. Precision represents the fraction of predicted positives that are correct, and recall indicates the proportion of actual positives identified. The F1-score is the harmonic measure that combines both precision and recall.} 
\end{table} 

\begin{sidewaystable}
\centering
\caption{Goodness-of-Prediction Indicators of Expanded Ground-Truth Questions for Benchmark -- Top 15 Model $\times$ Question Pairs}
\label{TABA2} 
\begin{tabular}{@{}llcccc@{}}
\\[-1.8ex]\hline \\[-1.8ex] 
\\[-1.8ex] Model & Ground-Truth & Accuracy & Precision & Recall & F1-Score \\ [1ex]
\hline \\[-1.8ex] 
GPT-4o (2024-11-20)      & Trust in political parties & 0.940 & 0.883 & 0.940 & 0.911 \\
GPT-4o-mini (2024-07-18) & Trust in political parties & 0.940 & 0.883 & 0.940 & 0.911 \\
Llama 4 Maverick (400B)  & Trust in political parties & 0.940 & 0.883 & 0.940 & 0.911 \\
GPT-4o-mini (2024-07-18) & Trust in Congress & 0.893 & 0.797 & 0.893 & 0.842 \\
GPT-4o (2024-11-20)      & Trust in Congress & 0.893 & 0.797 & 0.893 & 0.842 \\
Llama 4 Maverick (400B)  & Trust in Congress & 0.822 & 0.798 & 0.822 & 0.807 \\
Qwen 2.5 (32B)           & Trust in Congress & 0.777 & 0.797 & 0.777 & 0.779 \\
GPT-4o (2024-11-20)      & Trust in justice & 0.820 & 0.707 & 0.820 & 0.748 \\
Llama 4 Maverick (400B)  & Trust in television & 0.807 & 0.716 & 0.807 & 0.744 \\
GPT-4o (2024-11-20)      & Trust in television & 0.821 & 0.680 & 0.822 & 0.744 \\
GPT-4o (2024-11-20)      & Trust in government & 0.807 & 0.711 & 0.807 & 0.723 \\
Llama 4 Maverick (400B)  & Trust in government & 0.795 & 0.695 & 0.795 & 0.737 \\
GPT-4o-mini (2024-07-18) & Trust in government & 0.778 & 0.715 & 0.778 & 0.732 \\
Qwen 2.5 (32B)           & Trust in government & 0.758 & 0.679 & 0.758 & 0.712 \\
GPT-4o-mini (2024-07-18) & Trust in television & 0.696 & 0.708 & 0.696 & 0.686 \\
\bottomrule\\[-1.8ex] 
\end{tabular}
\\
{\footnotesize {\itshape Note.} Accuracy reports the proportion of synthetic samples aligned with the probabilistic reference sample. Precision represents the fraction of predicted positives that are correct, and recall indicates the proportion of actual positives identified. The F1-score is the harmonic measure that combines both precision and recall.} 
\end{sidewaystable}

\end{document}